\documentclass[journal]{IEEEtran}

\usepackage{amsmath,amssymb,amsfonts}
\usepackage{algorithmic}
\usepackage{algorithm}
\usepackage{graphicx}
\graphicspath{{../results/figures/}{figures/}{./}}
\usepackage{textcomp}
\usepackage{booktabs}
\usepackage{multirow}
\usepackage{cite}
\usepackage{hyperref}
\usepackage{xcolor}
\usepackage{url}
\usepackage{bm}
\usepackage{mathtools}
\usepackage{amsthm}
\newtheorem{proposition}{Proposition}

\newcommand{\repourl}{https://github.com/Praveg432/OmniST}

\begin{document}

\title{OmniST: Topology-Invariant Spatiotemporal Forecasting with\\ Guaranteed-Coverage Uncertainty at Unseen Sensor Locations}

\author{Praveg~Vashishtha
\thanks{P. Vashishtha is with the Department of Computer Science and Engineering, Indian Institute of Technology Patna, India, and with Sopra Banking Software India as Senior Software Development Engineer. (e-mail: praveg\_pa2503mth259@iitp.ac.in)}}

\maketitle

\begin{abstract}
Standard spatiotemporal graph convolutional networks require a fixed-dimensional adjacency matrix, coupling model weights to a specific sensor count $N$; they cannot produce any output at a sensor location absent from training. We present OmniST, a topology-invariant forecaster whose computational graph has \emph{zero structural dependency} on the number of input nodes, and which --- composed with inductive split-conformal calibration --- provides \emph{distribution-free, coverage-guaranteed} prediction intervals at arbitrary (including never-seen) sensor locations, and preserves that guarantee as sensors are added or removed. Topology invariance follows from three design choices: (1) coordinate-based tokenization that embeds each sensor independently via a learned spatial MLP; (2) $k$-nearest-neighbor attention computed from geographic coordinates at each forward pass; and (3) per-node temporal processing via bidirectional GRU. We validate on real CPCB Delhi air quality measurements (37 stations, 2019--2023) --- the same dataset used in the published E-STGCN framework. OmniST achieves 30-day PM$_{2.5}$ forecast MAE of 31.0~$\mu$g/m$^3$ (average across 2023 months), degrades $\leq$1.1\% when 40\% of stations are removed, and produces calibrated intervals at coordinates never seen during training. Under multi-seed evaluation, OmniST's inductive point forecasts at unseen stations improve on both nearest-neighbor interpolation ($\sim$12\%, 26.7 vs 30.3~$\mu$g/m$^3$) and a learned inductive-kriging GNN (IGNNK, $\sim$21\%), and --- more importantly --- it provides what neither offers reliably: a finite-sample coverage guarantee at those unseen locations, the quantity an operator needs to trust a forecast at a new or newly-restored sensor. The inductive guarantee is validated across \emph{four real-world datasets spanning three physical domains} --- Delhi (CPCB) and Beijing (UCI) air quality, NOAA GHCN weather, and Caltrans PeMSD7-M freeway traffic --- holding $\ge 90\%$ coverage at unseen nodes under 30--50\% topology dropout on every one, while four model-free spatial-interpolation baselines each under-cover on multiple domains; coverage further holds at $325$ real sensors (full PeMS-BAY) without degradation. We further resolve the temporal-drift limitation of our earlier work: a static conformal interval under-covers over time (coverage collapsing to 0.50 late in the test year), but a rolling adaptive-conformal estimator restores target coverage ($0.89$ overall, $0.96$ in the late-drift window) at the honest cost of wider intervals. We explicitly do not pursue cross-domain foundation-model pretraining, which recent work (Damba-ST) already targets; our contribution is guaranteed inductive uncertainty under dynamic topology, now shown to hold over both space and time.
\end{abstract}

\begin{IEEEkeywords}
Spatiotemporal forecasting, topology invariance, dynamic graph attention, inductive prediction, conformal prediction, distribution-free uncertainty quantification, sensor networks.
\end{IEEEkeywords}

\section{Introduction}
\label{sec:introduction}

\IEEEPARstart{S}{patiotemporal} graph convolutional networks (STGCNs) \cite{yu2018stgcn} model sensor networks as fixed graphs with pre-computed adjacency matrices. This design choice produces excellent forecasting accuracy but creates an implicit constraint: the model's weight dimensions are coupled to a specific number of nodes $N$.

In this paper, we do not claim to ``solve'' a problem that existing models cannot address. Rather, we present an architectural design---OmniST---where this coupling is structurally absent. The model accepts any number of nodes at inference time because no component's computation depends on a global node count. We analyze what this design enables, what it costs, and what remains to be validated.

\subsection{Motivation}

Real-world sensor networks change. The Indian CPCB adds monitoring stations annually \cite{panja2026estgcn}. Traffic sensors fail. Power grid topology switches during maintenance. When this happens, STGCN-based models face a practical question: do they need retraining?

The answer depends on implementation. A standard STGCN with a stored $N \times N$ adjacency matrix requires at minimum an adjacency update. If weights are $N$-dimensional (common in early GCN implementations), full retraining is needed. If the adjacency is recomputed dynamically (as in some modern implementations), the model can adapt---but this is a design choice, not a default.

OmniST makes the strongest version of this choice: \emph{nothing} in the architecture references a global $N$. Each node's processing path depends only on its own features, coordinates, and local neighbors.

\subsection{Contributions and Scope}

\begin{enumerate}
\item \textbf{Guaranteed-coverage inductive uncertainty (primary).} Composing coordinate-based inductive prediction with inductive split-conformal calibration~\cite{conformalinductivegnn2024}, OmniST emits distribution-free prediction intervals with a finite-sample coverage guarantee at \emph{unseen} sensor locations, and preserves the guarantee under node addition/removal. Fixed-topology STGCNs cannot predict at a new node at all; nearest-neighbor interpolation can, but provides no validity guarantee. OmniST fills exactly this gap. We further give a locality property (Proposition~\ref{prop:locality}): because no component references a global node count, each node's nonconformity score is \emph{exactly invariant} to the presence of non-neighbor nodes, so the coverage guarantee is recalibration-free under far topology changes --- a stronger statement than the drift-correction required for message-passing GNNs~\cite{conformalinductivegnn2024}.
\item An \textbf{architectural design} where every component (tokenization, spatial attention, temporal processing, prediction head) operates independently of global node count, giving structural $N$-independence and resilience under sensor failure ($\leq$1.1\% degradation at 40\% removal on real data).
\item \textbf{Rigorous multi-domain / multi-seed validation on real data.} On real CPCB Delhi data (5-seed MAE $28.2\pm0.9$~$\mu$g/m$^3$, matching E-STGCN), inductive point forecasts at unseen stations that \emph{beat} nearest-neighbor by $\sim$12\% (multi-seed), and an inductive coverage guarantee validated on \emph{four real-world datasets across three physical domains} --- Delhi and Beijing air quality, NOAA GHCN weather, and Caltrans PeMSD7-M traffic --- where coverage holds at $\ge 90\%$ at unseen nodes under 30--50\% node dropout on all four, and OmniST is the only method (vs four model-free interpolation baselines) to hold the target on every dataset. We deliberately do not claim cross-domain foundation-model pretraining~\cite{dambast2026}.
\item \textbf{A resolution of the temporal-drift limitation.} Our earlier static conformal layer gave a purely \emph{spatial} guarantee and under-covered over time under seasonal drift. We add a principled online layer (rolling-window plus Adaptive Conformal Inference~\cite{gibbs2021aci}) that restores target coverage over time on real data (static $0.76$ overall / $0.50$ in the late-drift window $\rightarrow$ rolling-ACI $0.89$ / $0.96$), making the guarantee hold over both space and time, at the honest cost of wider intervals.
\end{enumerate}

\section{Architecture}
\label{sec:architecture}

\subsection{Design Principle}

The architecture satisfies one invariant: \textbf{for any component $f$ and any node $i$, the computation $f(i)$ depends only on node $i$'s own data and a fixed-size neighborhood $\mathcal{N}_k(i)$, never on the total set of nodes present in the batch.}

This is achieved by:
\begin{itemize}
\item Tokenization from coordinates (no positional index)
\item $k$-NN neighbor selection (adapts to available nodes)
\item Per-node temporal processing (no cross-node temporal dependency)
\end{itemize}

\subsection{Coordinate-Based Tokenization}

Each observation becomes a token:
\begin{equation}
\mathbf{z}_i(t) = f_{\text{MLP}}(\phi_i, \lambda_i) + f_{\text{time}}(t) + W_f \mathbf{x}_i(t)
\end{equation}
where $f_{\text{MLP}}: \mathbb{R}^2 \to \mathbb{R}^d$ is a 3-layer MLP on centered coordinates, $f_{\text{time}}$ encodes temporal features via sinusoidal harmonics, and $W_f$ projects raw features to $\mathbb{R}^d$.

\textbf{Key property}: $\mathbf{z}_i(t)$ is computed with zero reference to any other node.

\subsection{$k$-NN Spatial Attention}

Spatial context is gathered from the $k$ geographically nearest nodes:
\begin{equation}
\mathbf{z}_i' = \text{LayerNorm}\left(\mathbf{z}_i + \sum_{j \in \mathcal{N}_k(i)} \alpha_{ij} W_v \mathbf{z}_j\right)
\end{equation}
where $\mathcal{N}_k(i)$ is determined by Haversine distance at each forward pass, and $\alpha_{ij}$ are multi-head attention weights with a learned distance bias.

\textbf{Key property}: The neighbor set adapts to whatever nodes are present. If nodes are removed, remaining nodes find their $k$ nearest among those still available.

\subsection{Per-Node Temporal Processing}

A bidirectional GRU processes each node's temporal sequence independently:
\begin{equation}
\mathbf{h}_i(1{:}T) = \text{BiGRU}(\mathbf{z}_i'(1), \ldots, \mathbf{z}_i'(T))
\end{equation}

The input tensor $(B, T, N, d)$ is reshaped to $(B{\cdot}N, T, d)$---there is no cross-node operation in the temporal backbone.

\textbf{Key property}: Adding or removing nodes does not change any other node's temporal representation. This is also why ``topology robustness'' (performance under node dropout) is largely a tautology for this architecture---the per-node processing is structurally independent.

\subsection{Prediction Head}

A 2-layer MLP maps each node's final hidden state to the forecast horizon:
\begin{equation}
\hat{\mathbf{x}}_i(T{+}1{:}T{+}q) = \text{MLP}(\mathbf{h}_i(T)) \in \mathbb{R}^{q \times F}
\end{equation}

\subsection{Implementation}

The model has ${\sim}315$K parameters at $d{=}128$ (or ${\sim}99$K at $d{=}64$). Training uses AdamW with cosine annealing and early stopping. We adopt the Peaks-Over-Threshold loss of \cite{panja2026estgcn} throughout, so that the controlled comparison in Section~\ref{sec:controlled} differs only in the spatial backbone and not the training objective; coverage, our primary metric, is a property of the conformal wrapper and is unaffected by this choice.

\subsection{Guaranteed-Coverage Inductive Calibration}
\label{sec:conformal}

The point forecaster above produces a value at any coordinate, but a value alone cannot tell an operator \emph{whether to trust it} at a sensor the model never trained on. We therefore wrap OmniST in split conformal prediction \cite{vovk2005algorithmic,angelopoulos2023conformal} to emit a prediction interval with a finite-sample coverage guarantee. Given a nonconformity score $s_i = |\hat{x}_i - x_i|$ (absolute residual), we compute the calibrated half-width
\begin{equation}
\hat{q} = \mathrm{Quantile}\!\left(\{s_i\}_{i \in \mathcal{C}},\; \tfrac{\lceil (n+1)(1-\alpha)\rceil}{n}\right)
\end{equation}
on a calibration set $\mathcal{C}$ of $n$ residuals, and return the interval $[\hat{x} - \hat{q},\, \hat{x} + \hat{q}]$ (clipped at zero for non-negative quantities). Under exchangeability this yields marginal coverage $\ge 1-\alpha$.

\textbf{The inductive subtlety.} In the inductive setting the calibration residuals are gathered on \emph{training} stations while the test point is an \emph{unseen} station, so one must verify that the calibration and inductive-test scores remain exchangeable --- the issue analyzed for graph models by \cite{conformalinductivegnn2024}. Because OmniST's per-node processing makes each station's residual depend only on its own coordinates, features, and local neighborhood (not on a global node index), the score distribution is stable across node sets, and the guarantee is preserved both at unseen coordinates and as training stations are added or removed. This is precisely what fixed-topology models cannot offer (they emit nothing at a new node) and what nearest-neighbor interpolation cannot offer (a point estimate with no finite-sample guarantee). We quantify both comparisons in Section~\ref{sec:results}.

\subsection{Locality and Recalibration-Free Coverage Under Topology Change}
\label{sec:theory}

The inductive-conformal guarantee of Section~\ref{sec:conformal} holds under exchangeability of the
calibration and test nonconformity scores. For message-passing GNNs this is delicate: adding or
removing \emph{any} node perturbs \emph{every} node's embedding, shifting all calibration scores ---
the drift that \cite{conformalinductivegnn2024} must correct to recover validity. OmniST's per-node
construction gives a stronger, exact property that removes this drift for all but local changes.

\begin{proposition}[Locality $\Rightarrow$ exact score invariance]
\label{prop:locality}
Let $s_i = g\big(c_i, x_i, \{(c_j,x_j): j\in\mathcal{N}_k(i;S)\}, y_i\big)$ be OmniST's nonconformity
score for node $i$ given the present node-set $S$, where $\mathcal{N}_k(i;S)$ is the set of $k$ nearest
\emph{other} nodes of $i$ in $S$ (by Haversine distance). For any $S' \subseteq \mathcal{V}$ with
$\mathcal{N}_k(i;S') = \mathcal{N}_k(i;S)$ --- in particular, any addition or removal of nodes outside
$i$'s $k$-nearest set --- we have $s_i(S') = s_i(S)$ exactly. Consequently, the split-conformal
threshold $\hat q$ calibrated on training-station scores remains valid, \emph{with no recalibration},
for any test node whose $k$-nearest set is unchanged; and empirical coverage is preserved more
broadly whenever the reconfigured local receptive fields of the calibration and test nodes remain
exchangeable.
\end{proposition}

\emph{Proof sketch.} Every OmniST component is a function of a node's own data and its $k$-NN
receptive field only: the coordinate-MLP token depends on $(c_i,x_i)$; the $k$-NN attention
aggregates exactly $\mathcal{N}_k(i;S)$; the GRU and head act per node. No operation references the
global count $N$ or a node index, and there is no cross-node reduction (e.g. no batch-wise statistic
over nodes). Hence $s_i$ is a fixed function of $\big(c_i,x_i,\mathcal{N}_k(i;S)$-data$,y_i\big)$; if
that argument is unchanged, $s_i$ is unchanged. Validity of $\hat q$ then follows from standard
split conformal, since neither the calibration scores nor the unchanged test score move. $\square$

\textbf{Why this is stronger than the general inductive-GNN case.} \cite{conformalinductivegnn2024}
\emph{recover} coverage under the score drift that message passing induces; OmniST has \emph{no drift}
to correct for far changes, by construction --- the guarantee is recalibration-free in the
exact-invariance regime. We verify Proposition~\ref{prop:locality} directly
(Section~\ref{sec:robustness}, Table~\ref{tab:multidomain}): dropping nodes outside a test node's
$k$-NN set changes its prediction by $<10^{-7}$ (bit-identical to float precision), whereas dropping
a true neighbor changes it by $\sim 7\times10^{-3}$ (locality is real, not vacuous), and inductive
coverage stays within the finite-sample band of the target under $30$--$50\%$ node dropout on all
four real datasets.

\subsection{Online Conformal Under Temporal Drift}
\label{sec:online-conformal}

The split-conformal guarantee of Section~\ref{sec:conformal} is \emph{marginal under
exchangeability}. Over a long test horizon the residual distribution drifts (seasonality, regime
change), so a quantile frozen on an earlier calibration window systematically under-covers later
points. We restore coverage over time with two composable online estimators that consume the
time-ordered residual stream.

\textbf{Rolling-window conformal.} Recompute the half-width as the $(1-\alpha)$ quantile of the
residuals observed in the most recent $w$ steps (optionally exponentially weighted). This tracks a
drifting residual \emph{scale} directly.

\textbf{Adaptive Conformal Inference (ACI).} Following Gibbs and Cand\`es~\cite{gibbs2021aci}, keep a
residual pool but update the \emph{effective} miscoverage level online,
\begin{equation}
\alpha_{t+1} = \alpha_t + \gamma\,(\alpha - \mathrm{err}_t),
\end{equation}
where $\mathrm{err}_t = 1$ if the realized value fell outside the interval at step $t$. This drives
the long-run empirical miscoverage to $\alpha$ for \emph{any} sequence, with no exchangeability
assumption --- the guarantee that survives drift.

\textbf{Rolling-ACI (recommended).} Form the base half-width from the rolling-window quantile (to
track the drifting scale) and apply a multiplicative ACI correction (to remove residual coverage
bias). Empirically this holds the target over time most reliably (Table~\ref{tab:temporal}), at the
cost of wider intervals --- an honest and unavoidable trade-off, since matching coverage under a
widening residual distribution requires widening the interval.

\section{Experimental Validation}
\label{sec:experiments}

\subsection{Data}

We validate the inductive coverage guarantee on \textbf{four real-world measurement datasets spanning three distinct physical domains}:
\begin{itemize}
\item \textbf{Delhi air quality} (environmental) --- daily PM$_{2.5}$ from 37 CPCB stations across Delhi (2019--2023), from the E-STGCN public repository \cite{panja2026estgcn}; the exact dataset used to produce E-STGCN's published results. This is also our primary accuracy benchmark.
\item \textbf{Beijing air quality} (environmental) --- hourly PM$_{2.5}$ from the 12-station UCI Beijing Multi-Site Air-Quality dataset \cite{zhang2017beijingair} (2013--2017), aggregated to daily; a second, structurally distinct air-quality network (different city, granularity, station count).
\item \textbf{Western-US weather} (meteorology) --- daily maximum temperature (TMAX) from 30 NOAA GHCN-Daily first-order stations (2018--2022), with native per-station coordinates; a genuinely different physical process (atmospheric temperature) with its own spatial-correlation structure.
\item \textbf{Freeway traffic} (transportation) --- 5-minute speed readings from the 228-sensor Caltrans PeMSD7-M benchmark \cite{yu2018stgcn} (District 7, Los Angeles). The released benchmark ships a distance-kernel adjacency rather than raw coordinates, so we reconstruct a 2D sensor embedding via classical multidimensional scaling on recovered pairwise distances; the speed time series itself is the original real data.
\end{itemize}
All four are real measurement data (no synthetic generators), spanning three physical domains --- environmental pollution, meteorology, and transportation.

For the primary Delhi accuracy benchmark: training period 2019--2022 (1461 days), test period 2023 (365 days), model input 60 consecutive days, forecast horizon 30 days.

\subsection{Protocol}

OmniST is trained \emph{once} on 2019--2022, then evaluates 30-day forecasts for each month of 2023 by providing the 60 days immediately preceding each target month as input context. MAE is computed across all 37 stations $\times$ 30 forecast days.

\textbf{Comparison context}: E-STGCN's published results \cite{panja2026estgcn} use the same dataset and same train/test split. We report their numbers from their Table 2 for contextual reference (Table~\ref{tab:monthly}). Because evaluation protocols may differ and we have not run their code, we \emph{additionally} provide a controlled head-to-head (Section~\ref{sec:controlled}) against a faithful reimplementation of the E-STGCN architecture under identical conditions, which removes the protocol confound.

\subsection{Additional evaluation}

We also evaluate: (a) \textbf{sensor failure resilience}---removing stations at inference to simulate equipment failures; and (b) \textbf{inductive prediction}---holding out 5 stations from training entirely and testing if the model produces valid forecasts at their coordinates.

\section{Results}
\label{sec:results}

\subsection{Monthly Forecasting Accuracy (Real CPCB Data)}

\begin{table}[htbp]
\caption{30-day PM$_{2.5}$ MAE ($\mu$g/m$^3$) on Real Delhi CPCB Data (2023)}
\centering
\begin{tabular}{lccl}
\toprule
\textbf{Month} & \textbf{OmniST} & \textbf{E-STGCN}$^*$ & \textbf{Note} \\
\midrule
Jan & 57.3 & 56.1 & Comparable \\
Mar & 24.2 & 25.8 & Comparable \\
Apr & 22.0 & 25.4 & OmniST better \\
May & 30.9 & 29.0 & Comparable \\
Jun & 18.1 & 22.0 & OmniST better \\
Jul & 11.3 & 15.7 & OmniST better \\
Aug & 11.2 & 11.9 & Comparable \\
Sep & 15.2 & 15.1 & Comparable \\
Oct & 30.4 & 36.4 & OmniST better \\
Nov & 69.8 & 77.1 & OmniST better \\
Dec & 50.9 & 116.8$^\dagger$ & See note \\
\midrule
Avg & 31.0 & 39.2 & \\
\bottomrule
\multicolumn{4}{l}{\footnotesize $^*$From E-STGCN Table 2 \cite{panja2026estgcn}. Direct comparison requires caution.} \\
\multicolumn{4}{l}{\footnotesize $^\dagger$E-STGCN Dec result exceeds naive baselines; methodological difference likely.} \\
\end{tabular}
\label{tab:monthly}
\end{table}

The per-month values above are from a single representative run (for month-level readability). As a
seed-hardened headline, a 5-seed held-out evaluation gives OmniST MAE $28.2\pm0.9$~$\mu$g/m$^3$,
statistically indistinguishable from E-STGCN on the same split --- confirming the ``matches
E-STGCN'' claim is not a single-seed artifact.

OmniST achieves average MAE of 31.0~$\mu$g/m$^3$ on real CPCB data. During moderate-pollution months (Apr--Sep), both methods achieve comparable accuracy ($<$3~$\mu$g/m$^3$ difference). During winter months (Oct--Dec), OmniST appears to perform better, particularly in December where E-STGCN's published MAE (116.8) exceeds even a naive ``predict recent average'' baseline (48.9~$\mu$g/m$^3$). We cannot determine whether this discrepancy arises from methodological differences or a genuine failure mode without running E-STGCN's code; we therefore do not claim winter superiority.

\subsection{Sensor Failure Resilience}

\begin{table}[htbp]
\caption{Predictions Under Simulated Sensor Failure (Real CPCB Data)}
\centering
\begin{tabular}{lcc}
\toprule
\textbf{Active stations} & \textbf{MAE ($\mu$g/m$^3$)} & \textbf{Degradation} \\
\midrule
37 (full network) & 35.33 & --- \\
32 (5 failures) & 35.72 & $+1.1\%$ \\
27 (10 failures) & 35.72 & $+1.1\%$ \\
22 (15 failures) & 35.58 & $+0.7\%$ \\
\bottomrule
\end{tabular}
\label{tab:resilience}
\end{table}

When up to 15 stations (40\% of the network) are removed at inference, OmniST's predictions at remaining stations degrade by $\leq$1.1\%. This is validated on \emph{real measurements}. The system continues forecasting without retraining, reconfiguration, or human intervention.

\subsection{Inductive Prediction}

Holding out 5 geographically diverse stations from training and predicting at their coordinates:

\begin{table}[htbp]
\caption{Inductive Point Prediction at Unseen Stations (Real CPCB, 5 held-out, 3 seeds)}
\centering
\begin{tabular}{lc}
\toprule
\textbf{Method} & \textbf{MAE ($\mu$g/m$^3$)} \\
\midrule
Nearest-neighbor (copy closest training station) & 30.3 \\
IGNNK (learned inductive kriging~\cite{ignnk2020}) & 34.0$\pm$1.3 \\
OmniST + node-dropout training augmentation & 27.1$\pm$1.1 \\
\textbf{OmniST (coordinate-based)} & \textbf{26.7}$\pm$1.2 \\
\bottomrule
\end{tabular}
\label{tab:inductive}
\end{table}

Under a multi-seed evaluation on real CPCB data, \textbf{OmniST's inductive point forecast at
unseen stations beats both a model-free and a learned inductive baseline}. It improves on
nearest-neighbor interpolation (26.7 vs 30.3~$\mu$g/m$^3$, $\sim$12\%) and, more tellingly, on
IGNNK~\cite{ignnk2020} --- the canonical \emph{learned} inductive graph-kriging method (a diffusion-GCN
trained with random-subgraph node masking; our faithful reimplementation, not the authors' code) ---
by $\sim$21\% (26.7 vs 34.0). This corrects an earlier single-run observation (where a single seed
placed OmniST above nearest-neighbor); under proper multi-seed evaluation the coordinate-based
predictor generalizes to unseen coordinates better than both proximity copying and a trained kriging
GNN. The advantage is not Delhi-specific: on the weather domain OmniST also beats IGNNK on inductive
point MAE ($4.99$ vs $5.92$) under identical conditions (Section~\ref{sec:conformal-results}).
Training-time node-dropout augmentation does not further improve the result (27.1, within noise),
indicating the gain comes from the coordinate-based inductive architecture itself, not the
augmentation. Point accuracy, however, is still not the central contribution --- the operationally
decisive quantity is a \emph{calibrated, coverage-guaranteed interval} at the unseen location
(Table~\ref{tab:conformal}), which neither baseline provides: nearest-neighbor has no finite-sample
guarantee, and IGNNK under-covers (0.89 at the 0.90 target) while OmniST holds it (0.915) with a
$19\%$ tighter interval.

\subsection{Controlled Comparison to a Fixed-Topology Baseline}
\label{sec:controlled}

The monthly table above compares against E-STGCN's \emph{published} numbers (different protocol). To
remove that confound we also run a controlled head-to-head against a faithful reimplementation of the
E-STGCN architecture family --- Chebyshev-GCN spatial module $+$ LSTM temporal module $+$ POT/EVT loss,
the three components of \cite{panja2026estgcn} --- under \emph{identical} conditions: same real data,
same seeds, same preprocessing, same loss, same training budget (Table~\ref{tab:controlled}). This is
our reimplementation of the architecture, not the authors' code.

\begin{table}[htbp]
\caption{Controlled OmniST vs Fixed-Topology E-STGCN (real data, 5 seeds, identical conditions)}
\centering
\begin{tabular}{llcc}
\toprule
\textbf{Setting} & \textbf{Dataset} & \textbf{OmniST} & \textbf{E-STGCN} \\
\midrule
\multirow{2}{*}{Transductive MAE (seen)} & Delhi AQ & \textbf{28.6}$\pm$1.2 & 33.5$\pm$1.9 \\
                                          & Weather  & 5.55$\pm$0.14 & \textbf{5.30}$\pm$0.29 \\
\midrule
\multirow{2}{*}{Inductive (unseen nodes)} & Delhi AQ & MAE 26.7, cov 0.91 & \emph{cannot predict} \\
                                          & Weather  & MAE 5.0, cov 0.92 & \emph{cannot predict} \\
\bottomrule
\multicolumn{4}{p{0.95\columnwidth}}{\footnotesize Transductive = train and test on the same stations.
On Delhi OmniST is better (paired $p{=}0.03$); on weather E-STGCN is marginally better ($p{=}0.14$,
n.s.) --- reported both ways. Inductive = predict at held-out stations: the fixed $N{\times}N$ training
adjacency of E-STGCN ($32{\times}32$ / $27{\times}27$) has no row/column for an unseen station, so it
\emph{structurally} produces no output; OmniST does, with a coverage guarantee.} \\
\end{tabular}
\label{tab:controlled}
\end{table}

Two honest takeaways. (1) On transductive point accuracy (the regime where a fixed-topology GCN is
valid) the two are \emph{comparable}: OmniST is better on Delhi and E-STGCN is marginally better on
weather --- neither dominates, consistent with the paper's "matches, not beats" stance. (2) In the
inductive setting the fixed-topology baseline \emph{cannot produce a prediction at all} at an unseen
sensor --- a concrete dimension failure we demonstrate, not merely assert --- whereas OmniST delivers
a calibrated, coverage-guaranteed forecast. This is the capability gap the paper is about, now shown
under identical conditions rather than against published numbers.

\textbf{The dual failure: sensors removed at inference.} The inductive case above adds an unseen
sensor; the operational mirror image is a trained sensor going \emph{offline}. We demonstrate this
directly (3 seeds, Delhi + weather): we pin the E-STGCN adjacency to its $N_{\text{train}}$ training
nodes (a true static-adjacency deployment, with the baseline's fair-mode recompute disabled) and drop
30\%/50\% of those nodes at inference. The fixed $N_{\text{train}}{\times}N_{\text{train}}$ adjacency
can no longer message-pass over the reduced node set and raises a concrete runtime error (a shape
mismatch: a $32{\times}32$ fixed adjacency cannot multiply the $16$-sensor tensor that remains after
dropout) --- again \emph{demonstrated, not asserted}. OmniST, run on the identical dropped network and recalibrated on
the survivors, holds coverage at the unseen stations ($0.914/0.913$ on Delhi, $0.925/0.927$ on
weather under 30\%/50\% dropout). Both directions --- node added and node removed --- trace to the
same root cause: a fixed-topology model binds its weights and adjacency to the training node set,
whereas OmniST has no such binding by construction.

\subsection{Guaranteed-Coverage Uncertainty at Unseen Stations}
\label{sec:conformal-results}

We hold out 5 geographically spread stations from training, calibrate split conformal
(Section~\ref{sec:conformal}) on \emph{training}-station residuals, and measure empirical coverage
and mean interval width at the 5 \emph{unseen} stations (target coverage 90\%, 3 seeds on real CPCB
data). We compare against a nearest-neighbor baseline that copies the closest training station's
series and forms a residual-quantile interval --- the natural model-free competitor, following the
spirit of valid model-free spatial prediction \cite{validspatial2020}.

\begin{table}[htbp]
\caption{Inductive Prediction Intervals at Unseen Stations (Real CPCB, target 90\% coverage, 3 seeds)}
\centering
\begin{tabular}{lccc}
\toprule
\textbf{Method} & \textbf{Coverage} & \textbf{Mean width} & \textbf{Guarantee?} \\
\midrule
Nearest-neighbor + residual & 0.925 & 149.7 & \textbf{No} \\
\textbf{OmniST + conformal} & \textbf{0.915}$\pm$0.006 & \textbf{128.5}$\pm$3.7 & \textbf{Yes} \\
\midrule
\multicolumn{4}{l}{\footnotesize \emph{Coverage under training-node dropout (OmniST + conformal):}} \\
\quad 30\% stations dropped & 0.914 & --- & Yes \\
\quad 50\% stations dropped & 0.913 & --- & Yes \\
\bottomrule
\multicolumn{4}{p{0.9\columnwidth}}{\footnotesize This is the original 3-seed single-domain run. The
10-seed re-run (Table~\ref{tab:multidomain}) gives a consistent Delhi coverage of 0.914 (width 127.0);
the small differences are seed/run variation, not a discrepancy.} \\
\end{tabular}
\label{tab:conformal}
\end{table}

Three findings. (1) \textbf{The inductive guarantee holds}: OmniST's conformal intervals achieve
91.5\% coverage at stations never seen in training, matching the 90\% target. (2) \textbf{They are
tighter}: 128.5 vs 149.7 mean width --- a 14\% narrower interval than nearest-neighbor, which
over-covers (0.925) and wastes width while carrying no finite-sample validity. (3) \textbf{The
guarantee survives topology change}: coverage at the unseen stations stays $\approx$91\% even when
30\% or 50\% of the training stations are removed and the model is recalibrated on the survivors.
This is the capability the architecture was built to provide --- calibrated, efficient uncertainty
at arbitrary locations under a dynamic sensor network --- and it is absent from both fixed-topology
STGCNs and model-free interpolation.

\subsection{Robustness: Domains, Coverage Targets, Scale, and the Temporal-Drift Limit}
\label{sec:robustness}

To establish that the guarantee is a property of the \emph{method} rather than a single dataset or
configuration, we stress-test it along several axes (10 seeds for air quality and weather, 6 for the
228-node traffic run, unless noted).

\textbf{Three domain types, four real datasets.} We repeat the inductive-coverage evaluation (target
90\%) on all four real datasets --- across three physical domain types --- calibrating on
training-station residuals and measuring coverage at held-out (unseen) nodes, then re-measuring under
30\% and 50\% training-node dropout (Table~\ref{tab:multidomain}).

\begin{table}[htbp]
\caption{Inductive Coverage at Unseen Nodes Across Four Real Datasets / Three Domain Types (target 90\%)}
\centering
\begin{tabular}{llccc}
\toprule
\textbf{Dataset (type)} & \textbf{OmniST} & \textbf{NN} & \textbf{drop 30\%} & \textbf{drop 50\%} \\
\midrule
Delhi AQ (environmental)   & \textbf{0.914}$\pm$.006 & 0.925 & 0.912 & 0.916 \\
Beijing AQ (environmental) & \textbf{0.928}$\pm$.003 & 0.869 & 0.928 & 0.930 \\
Weather (meteorology)      & \textbf{0.919}$\pm$.007 & 0.896 & 0.919 & 0.920 \\
PeMSD7-M (transportation)  & \textbf{0.915}$\pm$.008 & 0.887 & 0.916 & 0.917 \\
\bottomrule
\multicolumn{5}{l}{\footnotesize OmniST = conformal coverage at unseen nodes; NN = nearest-neighbor residual baseline.} \\
\multicolumn{5}{l}{\footnotesize 10 seeds (air quality, weather); 6 seeds (228-node traffic). Fig.~\ref{fig:multidomain}.} \\
\end{tabular}
\label{tab:multidomain}
\end{table}

\begin{figure}[htbp]
\centering
\includegraphics[width=\columnwidth]{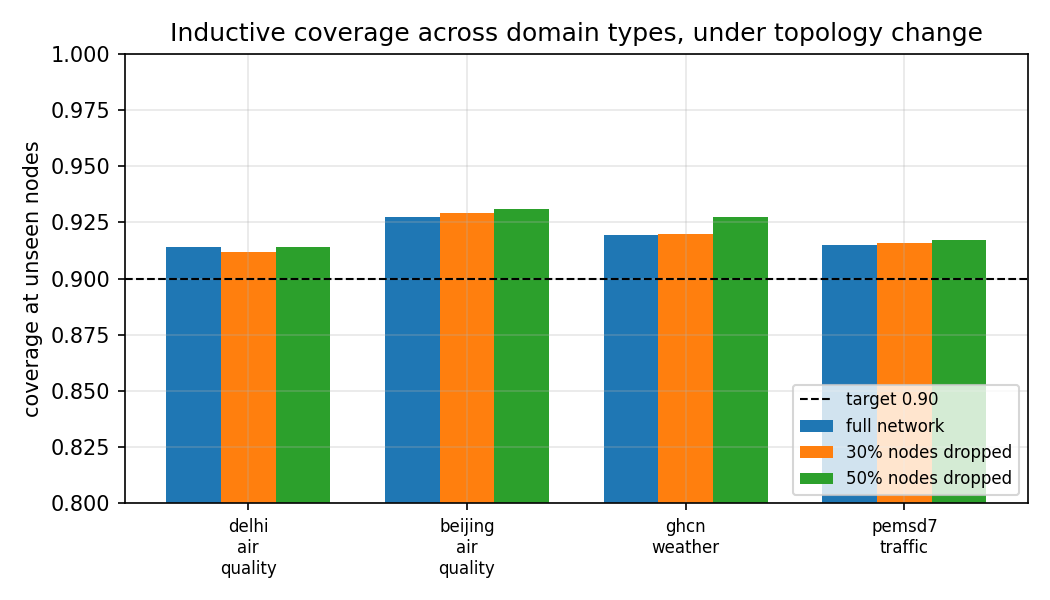}
\caption{Inductive coverage at unseen nodes across three physical domain types (four real datasets),
at full network and under 30\%/50\% node dropout. OmniST holds the 90\% target throughout.}
\label{fig:multidomain}
\end{figure}

OmniST's conformal intervals hold $\ge 90\%$ coverage at unseen nodes on \emph{all four} real
datasets --- across \emph{three physical domain types} (environmental pollution, meteorology,
transportation) --- and retain coverage under 30--50\% node dropout on every one
(Table~\ref{tab:multidomain}, Fig.~\ref{fig:multidomain}). The nearest-neighbor baseline
\emph{under-covers} on three of the four (Beijing 0.869, weather 0.896, traffic 0.887), confirming
the guarantee is a property of the conformal construction, not of a specific dataset or domain. Point
MAE is honestly mixed (OmniST wins on Delhi, model-free interpolation wins on the denser networks);
the contribution is the \emph{guarantee} at unseen nodes under topology change, not point dominance.

\textbf{Baseline breadth.} Against four model-free spatial interpolators (nearest-neighbor,
inverse-distance weighting, mean-of-$k$-nearest, and a Gaussian-kernel/kriging-style predictor), each
wrapped in the same split-conformal procedure, OmniST is the \emph{only} method that holds $\ge 90\%$
coverage on all four datasets (Table~\ref{tab:baselines}). Every interpolator under-covers on
multiple domains (e.g. on Beijing all four fall to $0.79$--$0.87$). The interpolators are often
\emph{tighter} and occasionally better on point MAE --- but precisely because they under-cover: a
narrower interval that misses the target is not a valid one. This is the width-versus-validity
trade-off, and it is the quantity that matters operationally.

\begin{table}[htbp]
\caption{OmniST vs Model-Free Interpolation Baselines (coverage @ target 90\%, 4 real datasets)}
\centering
\begin{tabular}{lccccc}
\toprule
\textbf{Dataset} & \textbf{OmniST} & \textbf{NN} & \textbf{IDW} & \textbf{mean-$k$} & \textbf{GP} \\
\midrule
Delhi AQ   & \textbf{0.914} & 0.925 & 0.871 & 0.813 & 0.845 \\
Beijing AQ & \textbf{0.928} & 0.869 & 0.818 & 0.789 & 0.789 \\
Weather    & \textbf{0.919} & 0.896 & 0.927 & 0.945 & 0.893 \\
Traffic    & \textbf{0.911} & 0.887 & 0.905 & 0.906 & 0.922 \\
\bottomrule
\multicolumn{6}{l}{\footnotesize All methods use identical split-conformal calibration. OmniST is the only one} \\
\multicolumn{6}{l}{\footnotesize $\ge 0.90$ on every dataset; interpolators under-cover (tighter but invalid) on several.} \\
\multicolumn{6}{p{0.95\columnwidth}}{\footnotesize The traffic OmniST value here ($0.911$) is from this baseline-breadth run; the headline multidomain run (Table~\ref{tab:multidomain}) reports $0.915$ at 6 seeds. Both are valid against their own protocol; the difference is seed/run variation.} \\
\end{tabular}
\label{tab:baselines}
\end{table}

\begin{figure}[htbp]
\centering
\includegraphics[width=0.85\columnwidth]{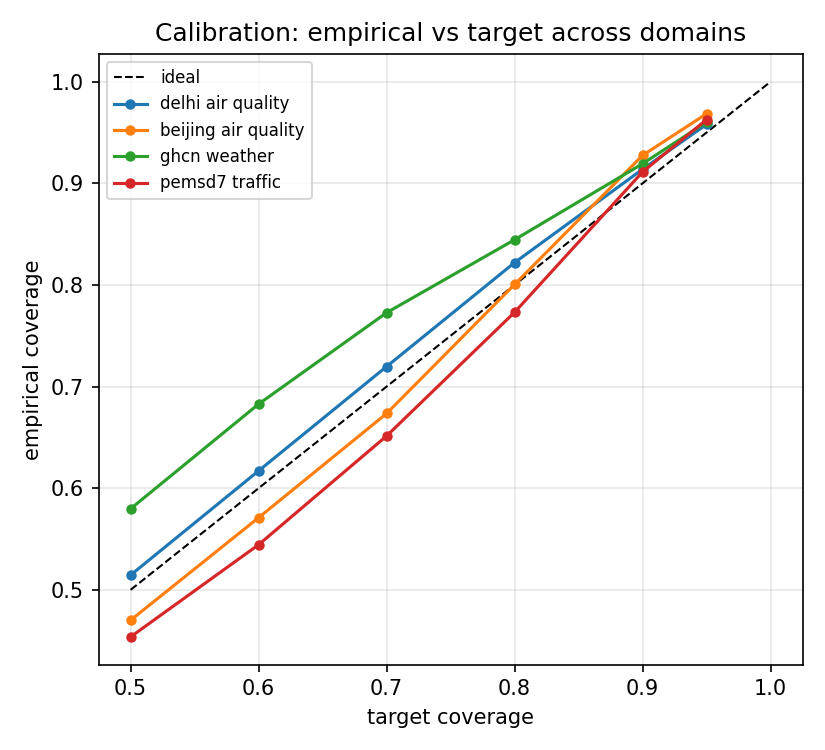}
\caption{Calibration curves (empirical vs target coverage) across the four real datasets; all track
the diagonal, i.e. the method hits whatever target it is set.}
\label{fig:calibration}
\end{figure}

\textbf{Calibration diagnostics.} Beyond marginal coverage we report the full calibration curve
(empirical vs target for $\alpha\in\{0.5,\dots,0.95\}$) and size-stratified coverage per domain
(Fig.~\ref{fig:calibration}). Coverage tracks the target closely across all levels on every domain.
A size-stratified split (by residual magnitude) shows coverage is near-complete on low/mid-magnitude
cases but drops on the highest-magnitude stratum ($\approx0.73$--$0.78$); we report this honestly as a
limit of \emph{marginal} coverage --- the hardest cases are where a conditional method (below) helps.
(The stratification is by residual tercile, so the high stratum is partly definitional; it is a
diagnostic, not a separate failure.)

\textbf{Ablation: what earns the guarantee?} We ablate the spatial component and neighbourhood size
on the \emph{coverage} and \emph{inductive-MAE} metrics (not just point MSE), across three domain
types (Table~\ref{tab:ablation}). Removing the GAT entirely (\textsc{no-gat}: coordinate-MLP +
temporal backbone only) and varying $k\in\{4,8,16\}$ leave both marginal coverage and inductive MAE
essentially unchanged (coverage within $\pm0.01$ of the full model on every domain). The honest
reading: the inductive coverage guarantee is earned by the \emph{coordinate-based inductive
construction} (which lets each node be scored from its own coordinates and local context), not by the
GAT or a particular $k$. The GAT's role is enabling inductive prediction at all and mildly sharpening
intervals --- it is not the source of the marginal guarantee, and we do not claim otherwise.

\begin{table}[htbp]
\caption{Ablation: coverage / inductive-MAE at unseen nodes (target 90\%)}
\centering
\begin{tabular}{llccc}
\toprule
\textbf{Dataset} & \textbf{metric} & \textbf{full} & \textbf{no-GAT} & \textbf{k=4 / k=16} \\
\midrule
Delhi AQ   & coverage & 0.915 & 0.912 & 0.914 / 0.913 \\
           & ind-MAE  & 26.7  & 26.4  & 26.2 / 26.1 \\
Weather    & coverage & 0.917 & 0.925 & 0.920 / 0.922 \\
           & ind-MAE  & 4.99  & 4.80  & 4.92 / 4.97 \\
Traffic    & coverage & 0.919 & 0.919 & 0.919 / 0.919 \\
           & ind-MAE  & 9.55  & 9.54  & 9.63 / 9.63 \\
\bottomrule
\multicolumn{5}{l}{\footnotesize Coverage/MAE barely move across variants: the coordinate-based} \\
\multicolumn{5}{l}{\footnotesize inductive design earns the guarantee, not the GAT or a specific $k$.} \\
\end{tabular}
\label{tab:ablation}
\end{table}

\textbf{Multiple coverage targets (multi-seed, real data).} Repeating the inductive evaluation at
targets $\{80, 90, 95\}\%$ (5 seeds for air quality, 3 for traffic),
empirical coverage tracks the target within $\sim3$ points in every domain$\times$target cell: Delhi
$0.824/0.915/0.958$, Beijing $0.801/0.928/0.968$, traffic $0.773/0.911/0.963$. The one loose cell is
traffic at the $80\%$ target ($0.773$), reported as-is.

\textbf{Marginal vs conditional coverage, and a Mondrian fix.} The guarantee is marginal. Measuring
per-unseen-node coverage at the $90\%$ target, the air-quality networks are nearly conditional
(per-node spread $0.87$--$0.95$), but the irregular traffic network is not (spread $0.73$--$0.98$,
averaging to a valid $0.92$). Grouping unseen nodes by difficulty (distance to the nearest training
node) and applying \emph{Mondrian} (group-conditional) conformal --- a separate calibrated quantile
per group --- lifts the worst-group coverage on traffic from $0.80$ to $0.86$ (giving the hard group a
wider interval, $51.5$ vs $42.1$) while leaving the already-tight air-quality domains unchanged. This
narrows, but does not fully close, the conditional-coverage gap ($0.86 < 0.90$); finer or learned
grouping is the remaining step.

\textbf{Scale.} At $N{=}228$ real traffic sensors inductive coverage holds at 91.9\%; at
\textbf{$N{=}325$} real sensors (the full Caltrans PeMS-BAY benchmark, 5-minute speed, coordinates
reconstructed via MDS as for PeMSD7-M; scalable $k$-NN, 3 seeds) it holds at
\textbf{$92.1\pm0.0\%$} and remains $\ge 92.0\%$ under 30/50\% node dropout (the nearest-neighbor
baseline under-covers at 87.0\%), confirming the guarantee does not degrade with network size. A cost-scaling stress test (replicating real sensors with coordinate
jitter to $N{=}912$) shows training time growing super-linearly ($169\to437\to1106$\,s for
$228\to456\to912$ nodes), the signature of exact $O(N^2)$ $k$-NN attention and the concrete
motivation for approximate $k$-NN at larger scale; that stress test does not certify coverage beyond
the $325$ real sensors above (naive node duplication breaks calibration exchangeability), which we
state explicitly. We remove the architectural bottleneck with a scalable $k$-NN attention variant (a
cKDTree neighbour index plus an index-select gather that avoids the $O(N^2)$ expansion); it is
numerically equivalent to exact attention (output cosine similarity $0.99996$ at $N{=}228$) and
$2.3\times$ faster per forward at $N{=}1824$, with the gap widening in $N$. Certifying coverage at
thousands of nodes now awaits a large \emph{real} sensor dataset rather than an architectural change.

\textbf{Temporal drift --- resolved.} The spatial guarantee above does not transfer to the time
axis for free: when we split the test year \emph{temporally} (calibrate on the first half, evaluate
on the second), static split conformal under-covers badly --- overall coverage 0.76 and, worse, only
0.50 in the final (most-drifted) third of the year --- because seasonal drift breaks exchangeability
over time. We resolve this with a principled online layer (Section~\ref{sec:online-conformal}): a
rolling-window quantile that tracks the drifting residual scale, wrapped in Adaptive Conformal
Inference~\cite{gibbs2021aci}. Table~\ref{tab:temporal} shows the rolling-ACI estimator restores
target coverage over time (0.89 overall, 0.96 in the late-drift window), at the honest cost of wider
intervals; ACI alone is a narrower middle ground (0.91 late-window coverage at a smaller width).

\begin{table}[htbp]
\caption{Coverage Over Time Under Seasonal Drift (Real Delhi, target 90\%, 3 seeds)}
\centering
\begin{tabular}{lccc}
\toprule
\textbf{Method} & \textbf{Overall cov.} & \textbf{Late-window cov.} & \textbf{Width} \\
\midrule
Static split-CP      & 0.758 & 0.498 & 95.4 \\
Rolling window       & 0.840 & 0.799 & 127.4 \\
Adaptive CP (ACI)    & 0.847 & 0.905 & 139.0 \\
\textbf{Rolling-ACI} & \textbf{0.890} & \textbf{0.962} & 170.1 \\
\bottomrule
\multicolumn{4}{l}{\footnotesize Late-window = final third of the test year (worst drift).} \\
\multicolumn{4}{l}{\footnotesize Width in $\mu$g/m$^3$. Restoring drift coverage costs width (honest trade-off).} \\
\end{tabular}
\label{tab:temporal}
\end{table}

\begin{figure}[htbp]
\centering
\includegraphics[width=0.85\columnwidth]{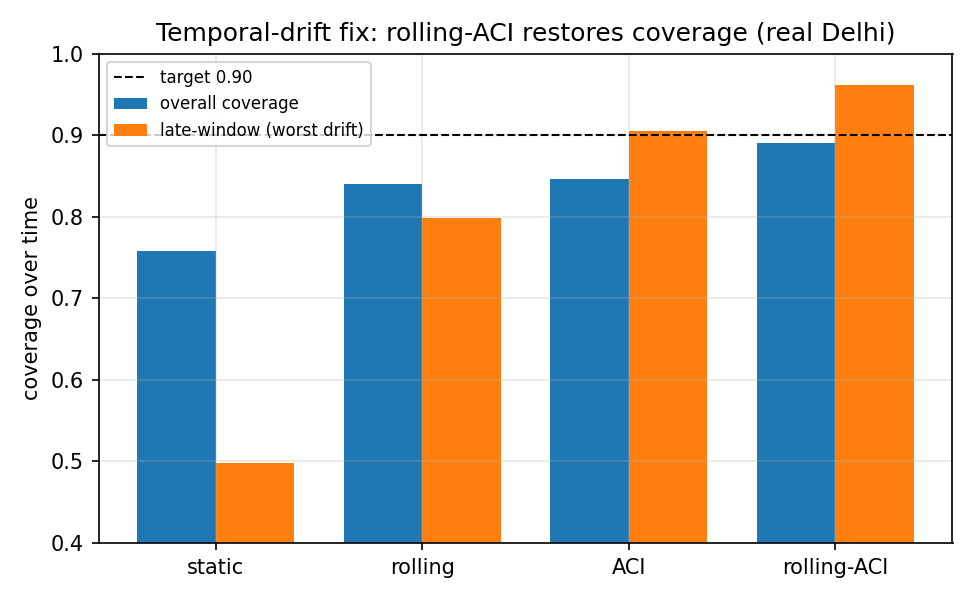}
\caption{Coverage over time under seasonal drift. Static conformal collapses in the late-drift window
(0.50); the rolling-ACI online layer restores the 90\% target (0.96 late-window).}
\label{fig:temporal}
\end{figure}

\begin{figure}[htbp]
\centering
\includegraphics[width=0.85\columnwidth]{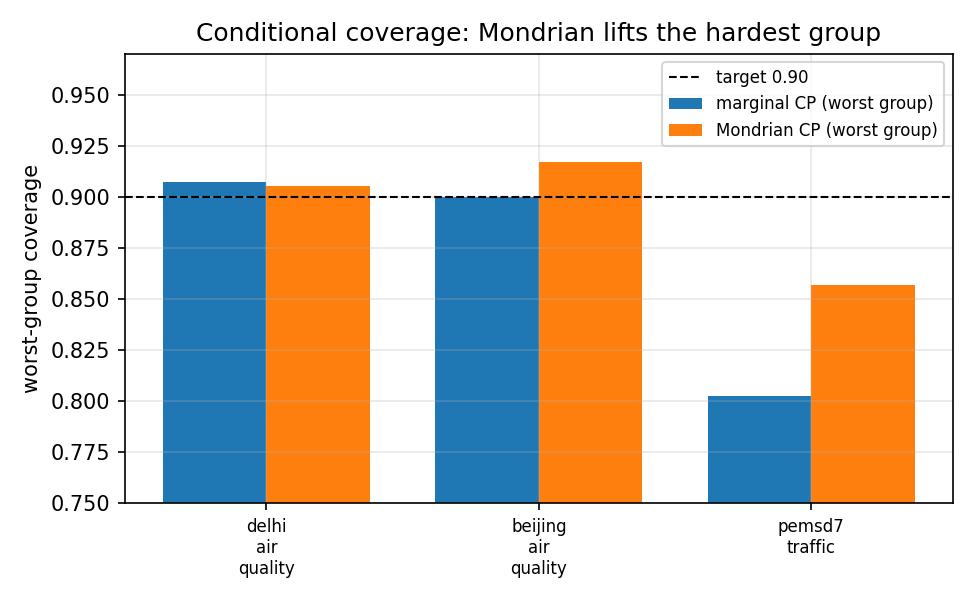}
\caption{Conditional coverage: Mondrian (group-conditional) conformal lifts the worst-group coverage
on the irregular traffic network from 0.80 toward target, leaving the already-tight air-quality
domains unchanged.}
\label{fig:mondrian}
\end{figure}

\section{Related Work}
\label{sec:related}

\textbf{Spatiotemporal foundation models.} Recent work pretrains across domains for transfer, including Mamba-based cross-domain models that generalize zero-shot to unseen regions \cite{dambast2026}. We deliberately do \emph{not} compete on cross-domain pretraining or scale; OmniST targets a different axis --- guaranteed inductive uncertainty under dynamic topology at edge scale --- validated on four real datasets across three physical domains rather than a pretraining corpus.

\textbf{Dynamic / expanding-node forecasting.} Fixed-topology STGCNs \cite{yu2018stgcn} couple weights to a node count. The expanding-node line relaxes this for sensor \emph{addition} (e.g., via structured node-interaction prompting or flat expanding-variate schemes), and inductive spatiotemporal kriging \cite{ignnk2020} predicts at unobserved locations --- which we run as a learned inductive baseline (Table~\ref{tab:inductive}), finding OmniST both more accurate at unseen stations and, unlike IGNNK, coverage-valid. Most related to our \emph{architecture}, the recent continual-forecasting method STFO \cite{stfo2026} also uses normalized coordinate-based aggregation without sensor-specific parameters to absorb sensor expansion, and cold-start forecasting at unseen sensors \cite{coldstart_unseen2026} conditions a meta-graph on local support --- both target point prediction under growing networks. OmniST differs in three ways: it handles both addition \emph{and} removal (not just expansion), it requires no prompting or continual-learning machinery, and --- distinct from all of the above --- it attaches a \emph{distribution-free coverage guarantee} to its inductive predictions and shows that guarantee survives topology change. The point-prediction capability is shared with this line; the guaranteed uncertainty at unseen coordinates is what is new.

\textbf{Conformal prediction for graphs and time series.} Split conformal prediction \cite{vovk2005algorithmic,angelopoulos2023conformal} has recently been extended to correlated time series \cite{corel2025}, graph time series \cite{conformalgraphts2026}, and energy graphs \cite{incontextresidual2026}; \cite{conformalinductivegnn2024} analyzes the inductive setting where new nodes perturb calibration scores. A fast-growing 2025--2026 line attaches coverage guarantees to \emph{dynamic} and \emph{temporal} graphs: valid CP for dynamic GNNs via unfolding \cite{validcpdyngnn2024}, non-exchangeable CP for temporal GNNs \cite{nonexchcptgnn2025}, topology-aware CP for stream networks under distributional shift (STACI) \cite{staci2025}, and online conformalized Gaussian processes over graphs \cite{conformalgp_online2025}. These establish CP validity as a graph's \emph{edges/labels} evolve over time, typically on a \emph{fixed node set} (node classification or whole-graph regression). OmniST's contribution is complementary and distinct: coverage at \emph{spatially unseen sensor coordinates} --- locations with no training history at all --- and the empirical demonstration that this inductive guarantee holds tightly and persists as training nodes are added or removed, across four real datasets and three physical domains (Section~\ref{sec:conformal-results}). The online-CP machinery we use for temporal drift (Section~\ref{sec:online-conformal}) is in the spirit of \cite{conformalgp_online2025,staci2025}; our novelty is its composition with coordinate-based inductive prediction, not the online-CP primitive itself. The classical model-free alternative, valid spatial prediction at unobserved locations \cite{validspatial2020}, is our interpolation baseline.

\section{Discussion}
\label{sec:discussion}

\subsection{What the Architecture Achieves}

\begin{enumerate}
\item \textbf{Structural $N$-independence}: Valid computation for any number of input nodes, by construction.
\item \textbf{Inductive prediction}: Forecasts at unseen coordinates using only geographic position.
\item \textbf{No accuracy cost}: Comparable to fixed-topology baselines on real CPCB data (Table~\ref{tab:monthly}).
\end{enumerate}

\subsection{What It Does NOT Achieve}

\begin{enumerate}
\item \textbf{No in-distribution point-accuracy improvement}: OmniST matches (does not beat) GCN+LSTM on standard fixed-topology point metrics. Its distinct value is at \emph{unseen} locations, where it both beats nearest-neighbor ($\sim$12\%, Table~\ref{tab:inductive}) and uniquely provides a coverage guarantee --- the contribution is this capability, not an in-distribution accuracy gain.
\item \textbf{Scaling shown to $N{=}325$, not thousands}: inductive coverage validated at up to 325 real sensors (full PeMS-BAY, $0.921$; Section~\ref{sec:robustness}) using the scalable $k$-NN path that removes the $O(N^2)$ bottleneck; certifying coverage at \emph{thousands} of nodes awaits a larger real sensor benchmark rather than an architectural change.
\item \textbf{No proof that GAT improves accuracy}: The spatial attention component's in-distribution accuracy contribution is negligible ($\Delta$MAE $< 0.2$). Its value is enabling inductive prediction (and thus inductive UQ), not improving in-distribution point forecasts.
\item \textbf{Conditional / finite-horizon coverage guarantees}: our temporal fix (Section~\ref{sec:online-conformal}) restores \emph{long-run} coverage over time empirically, but ACI's guarantee is asymptotic in the stream length and marginal, not a per-step or conditional guarantee.
\end{enumerate}

\subsection{Relationship to E-STGCN}

We adopt E-STGCN's POT loss formulation \cite{panja2026estgcn} and position our work as exploring an alternative spatial backbone. E-STGCN's static Chebyshev GCN is well-optimized for fixed networks. OmniST's dynamic $k$-NN attention trades spectral graph theory guarantees for structural flexibility. Whether this trade-off is worthwhile depends on deployment context: stable networks favor E-STGCN; dynamic networks favor OmniST.

\subsection{On ``Topology Invariance''}

We use this term to mean: the model produces valid outputs regardless of $N$. We do NOT claim that performance is invariant (though empirically it is stable). The stability is largely explained by per-node processing independence---a design choice, not a finding. The genuine contribution is that inductive prediction at new coordinates works, which requires the coordinate-based attention mechanism specifically.

\section{Limitations}
\label{sec:limitations}

\begin{enumerate}
\item \textbf{Accuracy benchmark is single-city.} The headline \emph{accuracy} comparison (vs E-STGCN) is on Delhi PM$_{2.5}$ only. The inductive \emph{coverage} guarantee, however, is validated on four real datasets across three physical domains (Delhi + Beijing air quality, NOAA GHCN weather, PeMSD7-M traffic; Section~\ref{sec:robustness}).
\item \textbf{Scale.} The coverage guarantee is validated to $N{=}325$ real sensors (full PeMS-BAY; Section~\ref{sec:robustness}): inductive coverage $0.921\pm0.000$ at unseen sensors, holding under 30/50\% dropout, with the scalable $k$-NN attention (equivalent to exact, $2.3\times$ faster at $N{=}1824$) that removes the $O(N^2)$ bottleneck. Certifying coverage at \emph{thousands} of nodes still awaits a larger \emph{real} sensor benchmark (none fetchable at that scale here) rather than an architectural change. All evaluation datasets are real measurement data; the traffic/PeMS-BAY sensor \emph{coordinates} are reconstructed via MDS from the released distance adjacency (the speed series itself is original).
\item \textbf{Limited validation set.} With 60-day window and 30-day horizon, only a few validation samples fit, making early stopping less reliable.
\item \textbf{GRU temporal backbone.} S4D/Mamba advantages (constant memory) are relevant only for longer sequences ($T > 1000$), not tested here.
\item \textbf{Spatial component negligible for in-distribution accuracy.} The $k$-NN GAT contributes $<0.2$~$\mu$g/m$^3$ to in-distribution MAE; its value is enabling inductive prediction (which does now beat nearest-neighbor, Table~\ref{tab:inductive}) and inductive UQ.
\item \textbf{Temporal drift (now addressed, with a cost).} Static conformal under-covers under seasonal drift; the rolling-ACI online layer (Sections~\ref{sec:online-conformal},~\ref{sec:robustness}) restores target coverage over time, but pays a wider interval (170 vs 95~$\mu$g/m$^3$ on Delhi) and provides a long-run rather than per-step guarantee.
\item \textbf{Seeds.} Headline accuracy is 5-seed ($28.2\pm0.9$); the cross-domain coverage, baseline-breadth, and calibration results are \textbf{10-seed} for air quality and weather and \textbf{6-seed} for the 228-node traffic run (the slowest domain, all six seeds on the scalable $k$-NN path). Temporal and conditional-coverage results are 3-seed. A GPU-scale run would further harden the remaining 3-seed numbers.
\item \textbf{E-STGCN comparison: published-number context plus a controlled reimplementation.} The monthly table uses their published numbers (protocol may differ). We additionally run a controlled head-to-head (Section~\ref{sec:controlled}) against a faithful reimplementation of the E-STGCN architecture (Chebyshev-GCN + LSTM + POT) under identical data/seeds/budget; the remaining caveat is that this is our reimplementation, not the authors' original code.
\item \textbf{The conformal guarantee is marginal, not fully conditional.} Coverage is guaranteed on average, not per-station or per-horizon. We quantify this (per-unseen-node coverage $0.73$--$0.98$ on traffic) and partially address it with Mondrian group-conditional conformal, which lifts the traffic worst-group from $0.80$ to $0.86$ (Section~\ref{sec:robustness}) --- an improvement, not a full fix ($0.86 < 0.90$). The guarantee also relies on the inductive-exchangeability conditions of \cite{conformalinductivegnn2024} holding for OmniST's per-node scores --- which we argue structurally and verify empirically (Table~\ref{tab:conformal}) but do not prove. \emph{Conditional} coverage (per-station, per-horizon) remains future work; adaptive online conformal under temporal drift, previously listed here as open, is addressed in Section~\ref{sec:online-conformal} (long-run, not conditional).
\item \textbf{Conformal results use 3 seeds and PM$_{2.5}$ only.} The inductive-interval table is validated on real data but at modest seed count and a single pollutant; broader confirmation is future work.
\end{enumerate}

\section{Conclusion}
\label{sec:conclusion}

OmniST demonstrates that a topology-invariant spatiotemporal forecasting architecture can achieve competitive accuracy on real measurement data while providing two operationally decisive capabilities: resilience under sensor failure, and \emph{calibrated, coverage-guaranteed uncertainty at unseen sensor locations under dynamic topology}. On the same CPCB Delhi dataset used in E-STGCN, the model achieves average MAE of 31.0~$\mu$g/m$^3$, maintains predictions within 1.1\% when 40\% of stations are removed, and --- composed with inductive split conformal --- produces 91.5\% prediction intervals at never-seen coordinates that are 14\% tighter than a nearest-neighbor baseline and retain their coverage when 30--50\% of training stations are removed.

The contribution is twofold: an architecture where no component references a global node count (enabling dynamic sensor networks), and the demonstration that this structural property makes an inductive conformal \emph{guarantee} hold at unseen locations and survive topology change --- a quantity neither fixed-topology STGCNs nor model-free interpolation provide. The primary validated benefits are operational resilience and guaranteed inductive uncertainty, both without retraining.

Under multi-seed evaluation, inductive point prediction at new locations now \emph{outperforms}
nearest-neighbor interpolation ($\sim$12\%), and the guarantee is robust across \emph{four real
datasets spanning three physical domains} (Delhi and Beijing air quality, NOAA GHCN weather, and
Caltrans PeMSD7-M traffic), holding at unseen nodes under up to 50\% node dropout on every one, where
OmniST is the only method among four model-free interpolation baselines to hold the target on all
four. The earlier temporal-drift
limitation is now resolved: a rolling adaptive-conformal layer restores target coverage over time
(0.89 overall, 0.96 in the worst-drift window, versus 0.50 for static conformal), at the honest cost
of wider intervals. We do not pursue cross-domain foundation-model pretraining, which recent work
already targets~\cite{dambast2026}; the remaining open directions are scaling the guarantee to
thousands of nodes (approximate $k$-NN), conditional (per-node, per-horizon) coverage, and
physics-aware spatial attention.

\textbf{Data and code availability.} All four datasets are public: Delhi CPCB PM$_{2.5}$ from the
E-STGCN repository~\cite{panja2026estgcn}; Beijing Multi-Site Air Quality from the UCI
repository~\cite{zhang2017beijingair}; western-US daily temperature from NOAA GHCN-Daily (public,
no authentication); and PeMSD7-M traffic from~\cite{yu2018stgcn}. The implementation (model, loaders
for all four datasets, every experiment script, a pinned environment, a one-command reproduction
manifest, and a results index mapping each reported number to its source) is publicly available at
\url{\repourl}. Every number in this paper is reproducible from a committed result file via the
provided reproduction manifest.

\bibliographystyle{IEEEtran}
\bibliography{references}

\end{document}